\documentclass[12pt]{article}
\usepackage[T1]{fontenc}
\usepackage[utf8]{inputenc} 
\usepackage{lmodern}  
\usepackage[ngerman,english]{babel}
\usepackage[a4paper,margin=2.9cm]{geometry}

\usepackage{csquotes}
\usepackage[
  backend=biber,
  style=authoryear,
  sorting=nyt,
  sortcites=true,
  maxbibnames=99
]{biblatex}

\DeclareNameAlias{sortname}{family-given}
\DeclareNameAlias{default}{family-given}

\usepackage{graphicx, amsmath, amsfonts, amssymb, mathtools, subcaption, multirow, makecell, authblk, float, comment, helvet, setspace, bbm, booktabs, tabularx, array, longtable, mathrsfs, bm, upgreek, hyperref, algorithm2e}

\title{Large language models as synthetic clinical experts to inform
longitudinal rare-disease modeling}

\author[1,2,*]{Clemens Schächter}
\author[3]{Astrid Pechmann}
\author[3]{Janbernd Kirschner}
\author[4,5]{Jan Hasenauer}
\author[1,2,6]{Harald Binder}

\affil[1]{Institute of Medical Biometry and Statistics (IMBI),
Faculty of Medicine and Medical Center -- University of Freiburg,
Freiburg, Germany}

\affil[2]{Freiburg Center for Data Analysis, Modeling and AI --
University of Freiburg, Freiburg, Germany}

\affil[3]{Department of Neuropediatrics and Muscle Disorders,
Faculty of Medicine and Medical Center -- University of Freiburg,
Freiburg, Germany}

\affil[4]{Bonn Center for Mathematical Life Sciences -- University of Bonn, Germany}

\affil[5]{Life and Medical Sciences (LIMES) Institute -- University of Bonn, Germany}

\affil[6]{CIBSS, Centre for Integrative Biological Signalling Studies --
University of Freiburg, Freiburg, Germany}

\affil[*]{Corresponding author}

\date{}

\begin{document}

\begin{titlepage}

\maketitle

\end{titlepage}

\section*{Summary}
\normalsize
\noindent 

Due to the limited amount of information, modeling longitudinal rare-disease data can particularly benefit from integrating clinical knowledge. 
Yet, elicitation of expert knowledge and formalization for model fitting is challenging, in particular due to limited time of clinical experts. 
To nevertheless make domain knowledge accessible during model fitting, we use large language models (LLMs) as synthetic clinical experts
to supervise a variational-autoencoder-based approach that learns low-dimensional latent summaries of visit-level observations.  
Specifically, LLMs are queried offline on textual descriptions of patient observations to obtain judgments, e.g., the suspected clinical category.
To improve the variational autoencoder fit, we train a differentiable surrogate model on these LLM-derived judgments and augment the loss function to encourage reconstructions that preserve the clinical-label distribution of their corresponding input profile.
In an application to real-world longitudinal motor-function assessments from children with spinal muscular atrophy in the SMArtCARE clinical registry, we map visit-level clinical profiles to low-dimensional representations that are linked by a multivariate mixed-effects model.
The synthetic expert loss discourages reconstructions that remain numerically close in data space but alter the clinical interpretation of the reconstructed motor function profile, 
such as by crossing a disease-type boundary. We thus reduced disagreement between original and reconstructed SMA type labels from about 11 to 7 percent. 
Furthermore, informing the latent representation by the synthetic expert improved prediction of motor function milestones as external endpoints compared with unsupervised latent representations and a data-level baseline. 
These results suggest that incorporating LLMs into model fitting can make clinical knowledge available to representation learning and improve clinical faithfulness for longitudinal rare-disease data.

\medskip
\noindent\textbf{Keywords:} Generative models; Large language models; Longitudinal modeling; Spinal muscular atrophy; Synthetic clinical experts; Variational autoencoders

\section{Introduction}
\label{introduction}

To incorporate clinical knowledge into statistical modeling, clinical experts need to devote time to interacting with modelers, 
and the latter must find ways to formalize the expert input into model structure or model fitting approaches. 
Clinical interpretation in rare-disease patient cohorts is often shaped by experience and contextual information to compensate for scarce data. 
Therefore, knowledge about distinctions between trajectories of patients that are meaningful in clinical practice may be hard to formalize as explicit formulas, priors, or constraints \parencite{Mikkola2023PriorElicitation,Daee2017KnowledgeElicitation}. 
Furthermore, the required sustained exchange between clinical and statistical experts makes the formalization time-consuming and vulnerable to errors arising from miscommunication across disciplines. 
As a result, statistical models for clinical data are often just optimized for objectives that are easier to formalize, 
such as the likelihood of data given a set of model parameters, prediction performance concerning some clinical endpoint, 
or reconstruction loss in representation learning. While these objectives are the established workhorses of statistical modeling, 
they can fail to preserve clinically relevant structure when that structure is not explicitly encoded in the optimization objective \parencite{Vickers2006decisionCurve,Kelly2019clinicalImpactAI,OakdenRayner2020HiddenStratification}. 

A novel approach to tackle this challenge could be offered by large language models (LLMs), 
which incorporate extensive biomedical knowledge from their training on clinical literature and can retrieve clinical information in natural language \parencite{Singhal2023ClinicalKnowledge,Thirunavukarasu2023llmMedicine}. 
Specifically, LLMs operationalize biomedical knowledge in text-continuation tasks, 
which has already been used successfully in medical explanation and question-answering tasks \parencite{Thirunavukarasu2023llmMedicine,Singhal2025medicalQA} and in clinical text summarization \parencite{VanVeen2024clinicalSummarization}. 
Incorporating question-answering tasks into model fitting provides an opportunity to make clinical knowledge available to statistical modeling in an automated, scalable manner. 
We specifically consider LLMs as synthetic clinical experts for generative models, namely conditional variational autoencoders (cVAEs) \parencite{VAE, cvae}, that provide dimension reduction for each time point where a patient has been observed.  
Here, LLM-derived judgments are used to encourage the cVAE to preserve clinical interpretations during reconstruction.

This is motivated by an application to a longitudinal dataset of children with spinal muscular atrophy (SMA), 
a rare neuromuscular disorder characterized by progressive muscle weakness and impaired motor-function development \parencite{Mercuri2022SMA,Schorling2020Advances}. 
Clinically, SMA severity is often summarized via three broad SMA type categories, which group patients by expected disease severity and motor milestone attainment \parencite{Calucho2018MotorMilestones,Mercuri2018DiagnosisManagement,Varone2025SMAClassification}. 
These categories support clinical description, define patient groups in studies, and inform clinical management and treatment planning \parencite{Varone2025SMAClassification,Schorling2020Advances,Mercuri2018DiagnosisManagement}. 
In our application, we focus on motor function of patients, which is measured using a specialized assessment \parencite{HINE,Bishop2018HINE2SMA} that records multiple ordinal test items covering a wide range of abilities such as head control, sitting, crawling, standing, and walking \parencite{HINE,Bishop2018HINE2SMA}. 
Since motor function profiles are interpreted relative to the child's age and expected attainment of developmental milestones, superficially similar profiles may receive different SMA type labels from a clinical expert \parencite{Calucho2018MotorMilestones,Bishop2018HINE2SMA}, 
such that no straightforward formalization is available, but an LLM might nevertheless be able to mimic expert assessment.

To model the development of motor function, we consider a VAE-based representation approach in which a cVAE learns visit-level latent variables and a multivariate mixed-effects model links these variables across observation times \parencite{Schaechter2025latentRepresentations,VAEMixedmodel2}. 
Dimension reduction by a cVAE helps to keep the number of parameters moderate in the face of a large number of motor function test items, 
which is particularly relevant in settings with small sample sizes, such as in a rare disease like SMA. 
However, a subsequent reconstruction from a latent variable may be numerically close to the observed profile while nevertheless crossing a clinically meaningful decision boundary.
Conversely, a numerically larger reconstruction error may be less important when the reconstructed profile retains the same functional interpretation.

To incorporate such potential differences in clinical meaningfulness into model fitting, 
LLMs can provide assessments of the original and reconstructed observations, and any discrepancies can inform the cVAE representation. 
For assessment by LLMs, we formulate pairwise question-answering tasks in which the LLM must assign a patient’s motor function profile to one of two candidate SMA type labels, 
separately for the original and reconstructed data. Performing these pairwise tasks over all combinations of SMA type labels allows us to identify inconsistencies in LLM answers 
before using them to guide the cVAE. To actually shape the cVAE representation, we need a differentiable criterion for augmenting the loss function. 
Therefore, we train an artificial neural network on the LLM judgments to obtain a differentiable surrogate of the synthetic expert that can provide gradients during VAE training. 

Our approach is related to work that uses LLM outputs not only as final predictions, but as supervision for subsequent learning procedures. 
In LLM-based weak supervision, prompted LLMs can be used to define labeling functions or generate weak labels that are then used to train downstream classifiers \parencite{Smith2022languageModelsLoop,Hsu2024KnowledgeFreeWeakSupervision}. 
Related distillation work uses LLM-generated labels and rationales as supervision for smaller task-specific models \parencite{Hsieh2023distillingStepByStep}. 
This is particularly relevant in clinical settings, where expert annotation is expensive and many judgments depend on contextual interpretation. 
However, such approaches usually use such labels to produce training data for classifiers or downstream predictors, rather than to steer the model fitting process of a statistical or generative model. 
Here, an additional challenge arises, because LLM judgments can be expensive, stochastic, and non-differentiable with respect to the parameters of the model that should be improved. 
Our use of a surrogate model for directly influencing model fitting follows a broader strategy in which complex, 
non-differentiable supervision is first converted into an auxiliary model before being used for optimization. 
Examples include knowledge distillation, where a student model is trained to approximate information provided by a larger teacher model \parencite{Hinton2015DistillingKnowledge,Hsieh2023distillingStepByStep}, 
and preference-based reinforcement learning, where comparison feedback is used to learn a reward model for policy optimization \parencite{Christiano2017DeepRLHumanPreferences,Ouyang2022TrainingLanguageModels}.  

This work extends LLM-based supervision beyond downstream label generation by using LLM-derived judgments directly in the fitting objective of a generative representation model to discourage clinically inconsistent reconstructions.
We investigate whether this supervision can improve the clinical faithfulness and milestone-prediction utility of latent representations learned from irregular longitudinal real-world SMA registry data.
Section~\ref{methods} formalizes our framework by introducing LLMs acting as synthetic clinical experts and then describing how their judgments are audited for internal consistency. These judgments are subsequently distilled into a differentiable surrogate model, and incorporated into VAE training to penalize reconstructions that remain numerically close to the input but alter the clinical interpretation of the profile. 
Section~\ref{results} evaluates the approach in longitudinal SMA motor-function data, where clinical categories are contextual, age-dependent, and difficult to encode as explicit rules, and examines whether synthetic-expert supervision reduces clinically inconsistent reconstructions and improves the prognostic usefulness of the resulting latent representations for motor milestone prediction. 
Section~\ref{discussion} considers methodological limitations and potential extensions.
\section{Methods}
\label{methods}

\subsection{Synthetic-expert label generation}
\label{synthetic_experts}

When clinical experts assess a clinical observation $\mathbf{y}\in\mathcal{Y}$ with context $\mathbf{s}\in\mathcal{S}$ and assign it to a clinical label $c_h\in\mathcal{C}=\{c_{1},\ldots,c_{H}\}$, 
they can follow fixed rules but may also rely on domain knowledge and experience to form their judgment. 
LLMs acting as synthetic clinical experts provide the possibility to retrieve such assessments based on domain knowledge in an automated manner and with short turnaround
time, when compared to real clinical experts. This opens a new way for incorporating knowledge in settings where fitted models are generative, i.e. provide means for sampling data from them: 
Instead of mathematically formalizing knowledge, e.g. as priors for parameter estimation, the LLM can be used to judge the sampled data in terms of clinical interpretability. 

Formally, LLMs work by segmenting input text into a sequence of tokens, which then are mapped to a high-dimensional space by an embedding model, 
resulting in a sequence of vectors $\mathbf{v}_\ell, \ell=1,\ldots,L$. 
These serve as the input for a transformer neural network $f_\mathrm{LLM}(\cdot)$, whose parameters were trained on a large corpus of text. 
While an LLM can generate arbitrary words as output, we constrain it (via corresponding instructions) to a forced-assignment task to one out of a pair of clinical labels.

This question format is related to paired-comparison approaches in psychometrics and to relative-similarity queries, 
where judgments are elicited by asking which of two alternatives better matches a reference case rather than by assigning an absolute category rating \parencite{Thurstone1927ComparativeJudgment,Tamuz2011CrowdKernel}.

Let $H$ denote the number of clinical categories in $\mathcal{C}$.
For each ordered pairwise combination of labels $m=1,\ldots,M=H(H-1)$, let $(a_m,b_m)$ with $a_m, b_m\in\{1,\ldots H\}, a_m\neq b_m$ denote the two label indices with corresponding candidate labels $c_{a_m}, c_{b_m}\in\mathcal{C}$.

For a rendered clinical profile and a specified pair of candidate labels, the LLM-induced pairwise judgment can then be written as
\begin{equation}
\begin{aligned}
c_{\mathrm{LLM}}^{(m)}
&= f_\mathrm{LLM}\left(\mathbf{v}^{\mathrm{instr},m}_1, \ldots, \mathbf{v}^{\mathrm{instr},m}_{L_1}, \mathbf{v}^{\mathrm{data}}_1, \ldots, \mathbf{v}^{\mathrm{data}}_{L_2}\right),\\
c_{\mathrm{LLM}}^{(m)}
&\in\{c_{a_m},c_{b_m}\}, \qquad m=1,\ldots,M.
\end{aligned}
\end{equation}
where $L_1$ of the $L$ input tokens correspond to the instructions (including specification of the respective pair $m$), and $L_2$ tokens correspond to the data $\mathbf{y}$ to be evaluated, with $L=L_1+L_2$.

For obtaining $\mathbf{v}^{\mathrm{instr},m}_\ell$, $\ell=1,\ldots,L_1$, and $\mathbf{v}^{\mathrm{data}}_\ell$, $\ell=1,\ldots,L_2$, from the embedding model, a decision is needed about what text to feed into the latter. 
The instruction text can be augmented with complementary visit information $\mathbf{s}$ that is relevant to form a clinical judgment, such as patient age or other covariates. 
To render the data $\mathbf{y}$ as text, each clinical variable can be paired with its observed value and a textual description of the measured quantity.
In some settings, numerical scores are ordinal codes whose meaning is defined by the clinical description rather than the numerical magnitude alone.
For example, in the SMA application, HINE-2 motor-function scores correspond to specific motor function abilities of the child.
In such cases, it may be preferable to provide the textual item descriptions instead of the raw numeric scores.

One advantage of the forced-assignment design is that the consistency of the LLM responses $c_{\mathrm{LLM}}^{(m)}, m=1,\ldots,M$ can be evaluated directly.
First, stability under label-order reversal can be assessed by comparing a pair of candidate label indices $(a_m,b_m)$ with its reversed counterpart $(b_m,a_m)$.
A response is considered stable if the same clinical label is selected in both directions.

Furthermore, when the clinical labels can be arranged in an ordinal order, the full set of pairwise answers can be checked for ordinal consistency.
Let $c_1 \prec c_2 \prec \cdots \prec c_H$ denote the clinical label order, for example
SMA type 1 $\prec$ SMA type 2 $\prec$ SMA type 3 $\prec$ presymptomatic motor development in our SMA application.
For a given profile, the pairwise answers are called ordinal consistent if there exists a single label index $k\in\{1,\ldots,H\}$ such that every pairwise choice is compatible with $c_k$ being the closest label on this ordered scale.
Formally, for each unordered comparison $m$ with $a_m<b_m$, choosing $c_{a_m}$ over $c_{b_m}$ imposes $k\leq (a_m+b_m)/2$, whereas choosing $c_{b_m}$ over $c_{a_m}$ imposes $k\geq (a_m+b_m)/2$.
If no such $k$ exists, the answer pattern is classified as internally inconsistent, whereas consistent answers are aggregated into a final synthetic-expert label $c_{\mathrm{SE}}=c_k\in\mathcal{C}$.
This ordinal-consistency check can be applied separately to the answers obtained under each candidate-label order.

\subsection{Synthetic-expert supervision}

We next describe how the synthetic-expert judgments enter the model-fitting objective.
For this, we consider a model $R_\omega$ with parameters $\omega$ that, for an observed clinical profile $\mathbf{y}\in\mathcal{Y}$ and context $\mathbf{s}\in\mathcal{S}$, produces a reconstruction
\[
\hat{\mathbf{y}} = R_\omega(\mathbf{y},\mathbf{s}) \in \mathcal{Y}, \text{ with } \hat{\mathbf{y}}\approx\mathbf{y}.
\]
This notation includes deterministic reconstruction maps as well as probabilistic generative models, in which $\hat{\mathbf{y}}$ denotes the decoder mean, predicted item probabilities, or a differentiable reconstruction summary used in the loss function.

We denote the synthetic-expert label of the observed profile by $c_{\mathrm{SE}}\in\mathcal{C}$ and the corresponding label of the reconstruction by $\hat{c}_{\mathrm{SE}}\in\mathcal{C}$.
We want to use the synthetic expert to steer the model $R_\omega$ toward reconstructions for which $\hat{c}_{\mathrm{SE}}$ aligns with $c_{\mathrm{SE}}$.
Since the parameters $\omega$ of the generative model $R_\omega$ are usually obtained by minimizing its loss function $\mathcal{L}_R(\omega;\mathbf{y},\hat{\mathbf{y}},\mathbf{s})$ via gradient-based methods, 
we require a differentiable approximation to the synthetic expert's class-score function.
Therefore, we train a surrogate neural network $S_\psi$ that approximates the synthetic-expert label given a sample $\mathbf{y}$ and contextual information $\mathbf{s}$
and maps this input to a probability distribution over the clinical labels:
\begin{equation}
    S_\psi(\mathbf{y},\mathbf{s})
=(\pi_1,\ldots,\pi_H)=\boldsymbol{\pi} \in\Delta^{H-1},
\qquad
\pi_h \approx \Pr(c_{\mathrm{SE}}=c_h\mid \mathbf{y},\mathbf{s}),
\quad h=1,\ldots,H,
\end{equation}
where $\pi_h$ approximates the probability that the LLM-derived final clinical label for profile $\mathbf{y}$ and contextual information $\mathbf{s}$ is $c_h$.
A sufficiently large training set for the surrogate can be generated by querying the LLM on real observations, on synthetic data sampled from a prefitted generative model, or on a combination of both. 

To measure inconsistency between the surrogate distribution for the observed profile $\boldsymbol{\pi}=S_\psi(\mathbf{y},\mathbf{s})\in\Delta^{H-1}$ and the corresponding distribution for the reconstruction $\hat{\boldsymbol{\pi}}=S_\psi(\hat{\mathbf{y}},\mathbf{s})\in\Delta^{H-1}$,
we calculate the Jensen-Shannon divergence,
\begin{equation}
    \operatorname{JS}\left(\boldsymbol{\pi}, \hat{\boldsymbol{\pi}}\right)
    =
    \frac{1}{2}\mathrm{KL}\left(\boldsymbol{\pi}\mid\mid\mathbf{m}\right)
    + \frac{1}{2}\mathrm{KL}\left(\hat{\boldsymbol{\pi}}\mid\mid\mathbf{m}\right),
    \qquad
    \mathbf{m}
    =
    \frac{1}{2}
    \left(
    \boldsymbol{\pi}+\hat{\boldsymbol{\pi}}
    \right).
    \end{equation}
where $\mathrm{KL}(\cdot\mid\mid\cdot)$ is the Kullback-Leibler divergence.

Finally, we include this term as a penalty in the loss function of $R_\omega$ to disincentivize reconstructed profiles for which the reconstruction label $\hat{c}_{\mathrm{SE}}$ differs from the original label $c_{\mathrm{SE}}$,
\begin{equation}
    \mathcal{L}_{R}^{\mathrm{SE}}(\omega;\mathbf{y},\hat{\mathbf{y}},\mathbf{s})=\mathcal{L}_R(\omega;\mathbf{y},\hat{\mathbf{y}},\mathbf{s})+\lambda_{\mathrm{SE}}\operatorname{JS}\left(\boldsymbol{\pi}, \hat{\boldsymbol{\pi}}\right).
\end{equation}
The scaling factor $\lambda_{\mathrm{SE}}$ weights the influence of the supervision term relative to the rest of the loss function.
During optimization of $R_\omega$, the surrogate parameters $\psi$ are held fixed, although gradients are propagated through the surrogate with respect to its reconstructed-profile input.

\subsection{Linking visit-level latent representations with a multivariate mixed-effects model}
\label{latent_mixed_model}

Having defined the synthetic-expert supervision signal, we next incorporate it into a representation model for longitudinal rare-disease data.

For this, let $i\in\mathscr{I}$ denote an individual for which longitudinal measurements $\mathbf{y}_{i,t}\in\mathbb{R}^n$ are documented at observation times $\mathscr{T}_i=\{t_{i,1},\ldots,t_{i,m_i}\}$, with $t_{i,1}<\cdots<t_{i,m_i}$. 
In rare-disease studies, the number of individuals may be small relative to the dimensionality and complexity of the visit-level measurements
Consequently, directly modeling high-dimensional visit-level observations with a statistical model such as a multivariate longitudinal mixed model can be infeasible.
We therefore use a conditional variational autoencoder (cVAE) \parencite{cvae,VAE} to represent each observation through a lower-dimensional latent variable $\mathbf{z}_{i,t}\in\mathbb{R}^d$, with $d<n$, such that statistical modeling becomes feasible. 
For this latent representation, we use the standard Gaussian prior
\[
p(\mathbf{z}_{i,t})=\mathcal{N}_d(\mathbf{0},\mathbf{I}_d).
\]
Because the synthetic-expert judgment depends on contextual information such as age, both encoding and decoding are conditioned on $\mathbf{s}_{i,t}$.
The encoder defines the approximate posterior
\[
q_{\boldsymbol{\phi}}(\mathbf{z}_{i,t}\mid \mathbf{y}_{i,t},\mathbf{s}_{i,t})
=
\mathcal{N}_d
\left(
\boldsymbol{\mu}_{i,t},
\operatorname{diag}(\boldsymbol{\sigma}_{i,t}^2)
\right),
\]
where $\boldsymbol{\mu}_{i,t}$ and $\boldsymbol{\sigma}_{i,t}$ are neural-network outputs.

To link longitudinal latent representations $\mathbf{Z}_i=(\mathbf{z}_{i,t})^\top_{t\in\mathscr{T}_i}\in\mathbb{R}^{m_i\times d}$, we follow our previous work \parencite{Schaechter2025latentRepresentations} and model them by multivariate mixed-effects regression,
\begin{equation}
\mathbf{Z}_i=\mathbf{X}_i\mathbf{B}+\mathbf{T}_i\mathbf{U}_i+\mathbf{E}_i.
\label{eq:latent_mixed_model}
\end{equation}
Here $\mathbf{X}_i\in\mathbb{R}^{m_i\times p}$ is the fixed-effect design matrix, $\mathbf{B}\in\mathbb{R}^{p\times d}$ contains population-level effects, 
$\mathbf{T}_i\in\mathbb{R}^{m_i\times q}$ is the random-effect design matrix, $\mathbf{U}_i\in\mathbb{R}^{q\times d}$ contains individual-specific deviations, 
and $\mathbf{E}_i\in\mathbb{R}^{m_i\times d}$  is residual error with
\begin{equation}
    \mathrm{vec}(\mathbf{U}_i)\sim
    \mathcal{N}_{qd}(\boldsymbol{0},\boldsymbol{\Phi}),
    \qquad
    \mathrm{vec}(\mathbf{E}_i)\sim
    \mathcal{N}_{m_id}(\boldsymbol{0},
    \boldsymbol{\Sigma}\otimes\mathbf{I}_{m_i}),
\end{equation}
where $\boldsymbol{\Phi}\in\mathbb{R}^{qd\times qd}$ and
$\boldsymbol{\Sigma}\in\mathbb{R}^{d\times d}$ are covariance matrices.
For example, in the SMA application, $\mathbf{X}_i$ may include visit age, genetic markers, treatment indicators and demographic baseline patient characteristics.
Similarly, $\mathbf{T}_i$ may include a patient-specific intercept and slope.

Given estimates $(\hat{\mathbf{B}}, \hat{\boldsymbol{\Phi}}, \hat{\boldsymbol{\Sigma}})$ and a mixed model prediction $\hat{\mathbf{z}}_{i,t}$ for $\mathbf{z}_{i,t}$, the decoder defines the conditional reconstruction distribution
\[
p_{\boldsymbol{\theta}}
(\mathbf{y}_{i,t}\mid \hat{\mathbf{z}}_{i,t},\mathbf{s}_{i,t}),
\]
with the distributional form chosen to match the clinical scale of each component of $\mathbf{y}_{i,t}$.

The approach is fitted using an estimation scheme in which the encoder and decoder parameters $(\boldsymbol{\phi}, \boldsymbol{\theta})$ are updated alternately with the latent mixed-effects model parameters $(\mathbf{B}, \boldsymbol{\Phi}, \boldsymbol{\Sigma})$.
First, the encoder and decoder parameters are updated while the latent mixed-effects model parameters are frozen.

While updating the encoder and decoder, we use the surrogate network to encourage consistency between observed and reconstructed profiles.
Specifically, we optimize a $\beta$-VAE-style objective \parencite{betaVAE} augmented by two penalties: one aligning the encoder representation with the latent mixed-model prediction, and one aligning the surrogate distributions of the observed and reconstructed profiles:
\begin{equation}
    \label{eq:synt_loss}
    \begin{aligned}
    \mathcal{L}_{\mathrm{cVAE-MM}}^{\mathrm{SE}}\big(\boldsymbol{\phi}, \boldsymbol{\theta}\big)
    =&\sum_{i\in\mathscr{M}}\frac{1}{|\mathscr{T}_i|}\sum_{t\in \mathscr{T}_{i}}\bigg(
    -\mathbb{E}_{\mathbf{z}_{i,t}\sim q_{\boldsymbol{\phi}}(\mathbf{z}_{i,t}\mid \mathbf{y}_{i,t},\mathbf{s}_{i,t})}
    \left[\log p_{\boldsymbol{\theta}}(\mathbf{y}_{i,t}\mid\hat{\mathbf{z}}_{i,t},\mathbf{s}_{i,t})\right]\\
    &+\beta \mathrm{KL}\left[q_{\boldsymbol{\phi}}(\mathbf{z}_{i,t}\mid \mathbf{y}_{i,t},\mathbf{s}_{i,t})\mid\mid p(\mathbf{z}_{i,t})\right]
    +\eta\|\hat{\mathbf{z}}_{i,t}-\mathbf{z}_{i,t}\|^2_2\\
    &+\lambda_{\mathrm{SE}}
    \operatorname{JS}
    \left(
    S_\psi(\mathbf{y}_{i,t},\mathbf{s}_{i,t}),
    S_\psi(\hat{\mathbf{y}}_{i,t},\mathbf{s}_{i,t})
    \right)
    \bigg),
    \end{aligned}
\end{equation}
where $\mathscr{M}\subset\mathscr{I}$ is a mini-batch. We define the reconstruction $\hat{\mathbf{y}}_{i,t}$ as the conditional mean of the decoder distribution 
$p_{\boldsymbol{\theta}}(\cdot\mid \hat{\mathbf{z}}_{i,t},\mathbf{s}_{i,t})$, rather than as a sample drawn from this distribution. 
This keeps the surrogate-based supervision term differentiable during model fitting.
After some iterations, the encoder and decoder parameters are frozen and latent representations are obtained for all observations. 
Subsequently, the latent mixed-model parameters $(\mathbf{B}, \boldsymbol{\Phi}, \boldsymbol{\Sigma})$ are estimated via maximum likelihood. 
These two optimization steps are alternated for a prespecified number of epochs or until a stated convergence criterion is met.

After the training process has concluded, the latent variables $\mathbf{z}_{i,t}$ provide low-dimensional summaries of the visit-level observations $\mathbf{y}_{i,t}$.
Using them in downstream tasks, such as time-to-event modeling, provides a way to assess whether the learned variables retain information relevant for predicting future events.
Specifically, for an event time $T_i$ and censoring indicator $\delta_i$, the observations of patient $i$ can be represented as time-varying intervals on the observation-time scale.
On an interval $(t_{i,j},t_{i,j+1}]$, a Cox model on the latent representation can be written as
\begin{equation}
\lambda_i(t\mid \mathbf{z}_{i,t_{i,j}})
=
\lambda_0(t)
\exp\left(\boldsymbol{\beta}^{\top}\mathbf{z}_{i,t_{i,j}}\right),
\qquad
t\in(t_{i,j},t_{i,j+1}] .
\end{equation}
Here, $\lambda_0(t)$ denotes the baseline hazard and $\boldsymbol{\beta}$ quantifies the association between the learned latent representation and the event hazard.
In the SMA application, this Cox model is used to evaluate whether the latent representation contains information about future attainment of motor-function milestones, such as sitting, standing, and walking.

\begin{figure}[!p]
    \includegraphics[width=\textwidth]{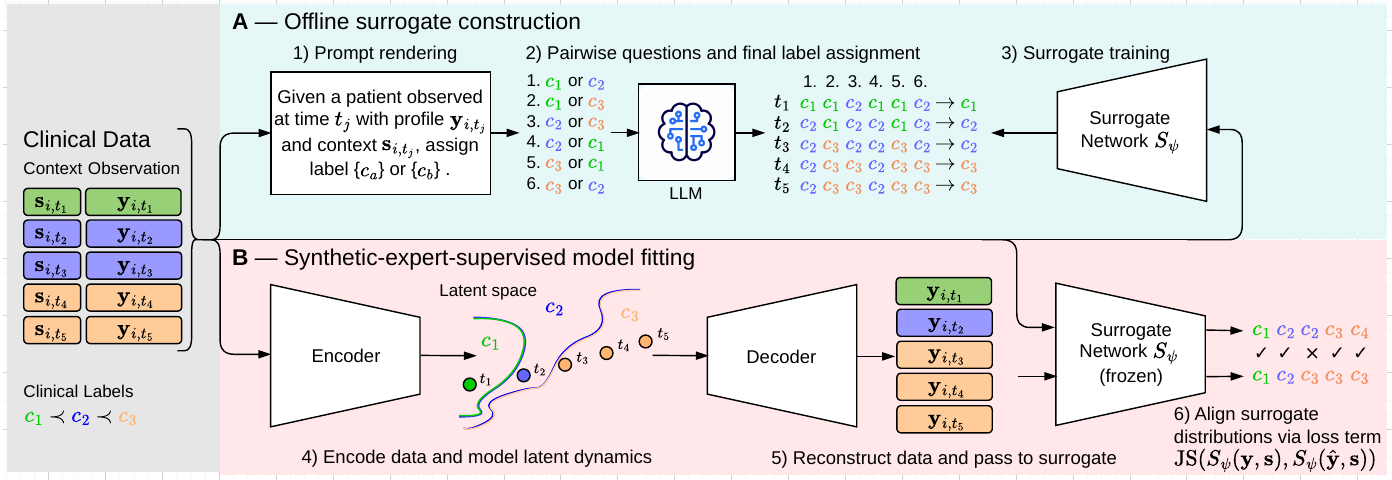}
    \caption{Overview of the proposed synthetic-expert supervision framework. \\
    Part A: Offline surrogate construction.
    1) Each observed clinical profile $\mathbf{y}_{i,t}$ and its contextual information $\mathbf{s}_{i,t}$ are rendered as a natural-language prompt.
    2) The LLM rater produces pairwise responses for the specified candidate labels, which are checked for consistency and aggregated into retained synthetic-expert labels. 
    3) The retained LLM-derived labels are used to train a differentiable surrogate network $S_{\psi}$ that maps clinical profiles and contextual information to distributions over the clinical labels. \\
    Part B: Synthetic-expert-supervised model fitting.
    4) The encoder maps the longitudinal observations to low-dimensional latent representations, whose trajectories are linked through the latent mixed-effects model.
    5) The decoder reconstructs the clinical profiles, and both the observed profiles $\mathbf{y}_{i,t}$ and their reconstructions $\hat{\mathbf{y}}_{i,t}$ are evaluated by the frozen surrogate network using the same contextual information $\mathbf{s}_{i,t}$. The third observation is reconstructed with the wrong label.
    6) The Jensen--Shannon divergence between the corresponding surrogate distributions is added to the training objective as a synthetic-expert loss term, thereby discouraging reconstructions whose clinical interpretation differs from that of the observed profile.}
    \label{fig:synthetic_expert_supervision}

    \par\smallskip
    \noindent\textbf{Alt text:}
    Graphical approach visualization divided into two stages. Part A shows clinical observations
    and contextual information being converted into text prompts. An LLM
    performs pairwise comparisons between clinical labels, inconsistent
    responses are identified, and the retained labels are used to train a
    differentiable surrogate network. Part B shows an encoder, longitudinal
    mixed-effects model, and decoder producing reconstructed clinical
    profiles. A frozen surrogate evaluates the original and reconstructed
    profiles, and the Jensen--Shannon divergence between their predicted
    label distributions is added to the training loss.
        
\end{figure}

Figure~\ref{fig:synthetic_expert_supervision} provides a graphical summary of the proposed workflow.
It highlights the separation between offline LLM-based label generation and the differentiable training loop, 
where the surrogate provides the path through which synthetic-expert feedback enters the cVAE loss.

\section{Results}
\label{results}

\subsection{Application to spinal muscular atrophy data}

We evaluated synthetic-expert supervision using longitudinal motor-function assessments from the SMArtCARE registry \parencite{SMArtCARE}, 
a clinical registry for patients with spinal muscular atrophy (SMA). 
In the SMArtCARE registry, disease severity is monitored through motor-function assessments at repeated follow-up visits using instruments such as the HINE-2 motor-function test \parencite{HINE,Bishop2018HINE2SMA}.
During these assessments, patients perform several motor-function tasks that are graded by physicians on ordinal item scales. 
These test results should not be interpreted in isolation: their clinical meaning depends on the child's age and developmental context
and similar HINE-2 profiles can therefore carry different clinical implications.

To model disease progression for these visit-level profiles, we used the cVAE-based representation approach introduced in Section~\ref{latent_mixed_model}. 
The cVAE compresses each HINE-2 profile, conditioned on the corresponding age value, into a low-dimensional latent representation and reconstructs the original item profile from this representation, 
while the longitudinal multivariate mixed-effects model links the latent variables across observation times.

For the synthetic-expert task, we used the three conventional symptomatic SMA type categories, SMA types 1--3, which reflect typical age at onset and expected motor-function progression \parencite{Mercuri2018DiagnosisManagement}. 
SMA type 1 typically begins in early infancy and is associated with severe muscle weakness and failure to achieve independent sitting. 
SMA type 2 usually describes patients who achieve sitting but not independent walking. 
SMA type 3 has symptom onset after infancy, and patients usually acquire independent walking for at least part of the disease course. 
We additionally included presymptomatic motor development as a fourth category for observations from patients who had been diagnosed with SMA but did not yet show clear clinical signs.

We used synthetic-expert supervision to disincentivize decoder reconstructions that reproduce the HINE-2 profiles accurately while changing the clinical reading of the profile in terms of these categories,
for example by shifting an age-appropriate pattern toward delayed motor development or by moving a profile slightly across an SMA type boundary. 

Specific implementation details for preprocessing, synthetic-expert generation, surrogate training, and model fitting are provided in Appendix A.

\subsection{Obtaining a differentiable synthetic-expert surrogate on SMA data}

We first assessed whether LLM-derived judgments on SMA motor-function profiles could be turned into a reliable differentiable supervision signal.
For this purpose, we queried LLMs from the Qwen-3 \parencite{Yang2025Qwen3}, GPT-OSS \parencite{OpenAI2025GPTOSS}, and Gemma-4 \parencite{GoogleDeepMind2026Gemma4} model families to classify textual descriptions of HINE-2 profiles paired with the patient's age. 
Each profile was evaluated through pairwise comparisons between the four labels, SMA types 1-3 and presymptomatic motor development.
The pairwise responses were then mapped to a final SMA type label by considering the ordinal ordering of the SMA type categories.

We used three criteria to evaluate consistency of the LLM answers. 
First, we used the pairwise comparison design to investigate ordinal consistency by counting contradictory patterns.
For example, when an LLM selects "SMA type 1" over "SMA type 2", but also "SMA type 3" over "SMA type 2" for the same profile, 
the resulting pattern is classified as inconsistent because it violates the ordinal structure of the SMA type labels. 
Second, we evaluated robustness to prompt wording by reversing the order in which the candidate labels were presented to the LLM. 
Finally, we queried the LLM to return an uncertainty score for each answer on a low, medium, or high scale and examined whether uncertainty was elevated for samples in which the LLM failed to recover the same label after label-order reversal.

Table~\ref{tab:llm_output_quality} summarizes the internal-consistency metrics for the evaluated LLM raters and for a baseline that independently selects either candidate with probability $p=0.5$.
Small models, such as Qwen-3-0.6B, showed output patterns that were close to random baseline, indicating that their pairwise label choices contained little stable clinical structure. 
With increasing model size, the consistency of the synthetic-expert judgments improved noticeably. 
Gemma-4-31B had the highest overall internal-consistency and order-stability metrics and yielded consistent final labels for around 89.9\% of samples and agreement under label-order reversal for 97.8\% of pairwise comparisons and 92.7\% of final SMA type labels. 
Notably, the stronger models also assigned higher self-reported uncertainty to answers that were not stable under label-order reversal. 
This effect was not apparent for smaller models such as Qwen-3-0.6B and Qwen-3-1.7B, but was $9.6$ times higher for Gemma-4-31B among order-unstable comparisons than among stable comparisons.

Subsequently, the surrogate model was trained to reproduce the final class label from age and the thermometer-encoded HINE-2 profile with two heads to reproduce both label-order choices. 
Profiles with contradictory pairwise decision patterns were excluded from the training set allowing the surrogate to interpolate for samples with inconsistent pairwise answers.
For the largest model in each family, surrogate accuracy exceeded $90\%$ in held-out folds in fivefold cross-validation (Table~\ref{tab:llm_output_quality}).

\begin{figure}[!p]
    \begin{subfigure}{\textwidth}
        \centering
        \includegraphics[width=\textwidth]{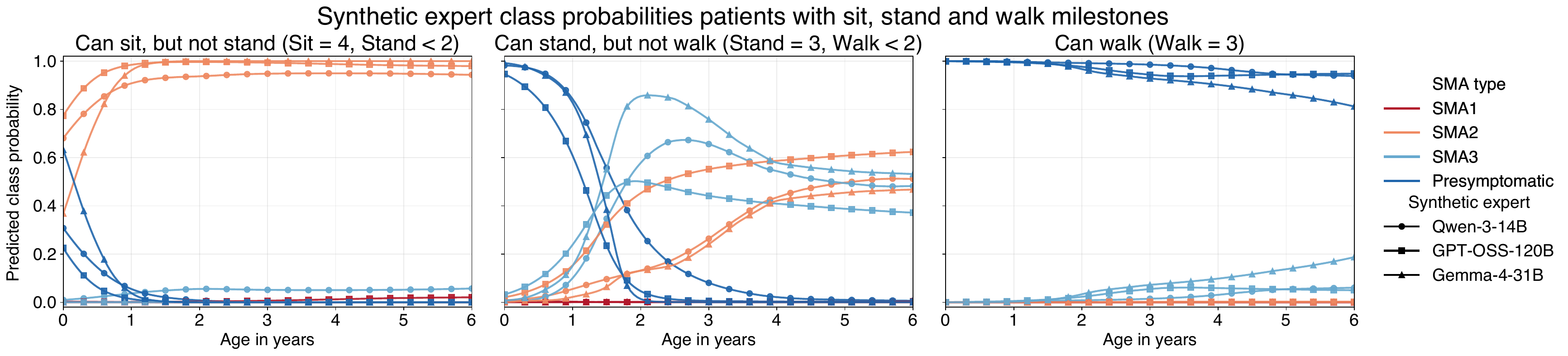}
        \caption{Age-dependent surrogate class probabilities for selected motor-function profiles.}
    \end{subfigure}
    \vspace{0.4em}
    \begin{subfigure}{\textwidth}
        \centering
        \includegraphics[width=\textwidth]{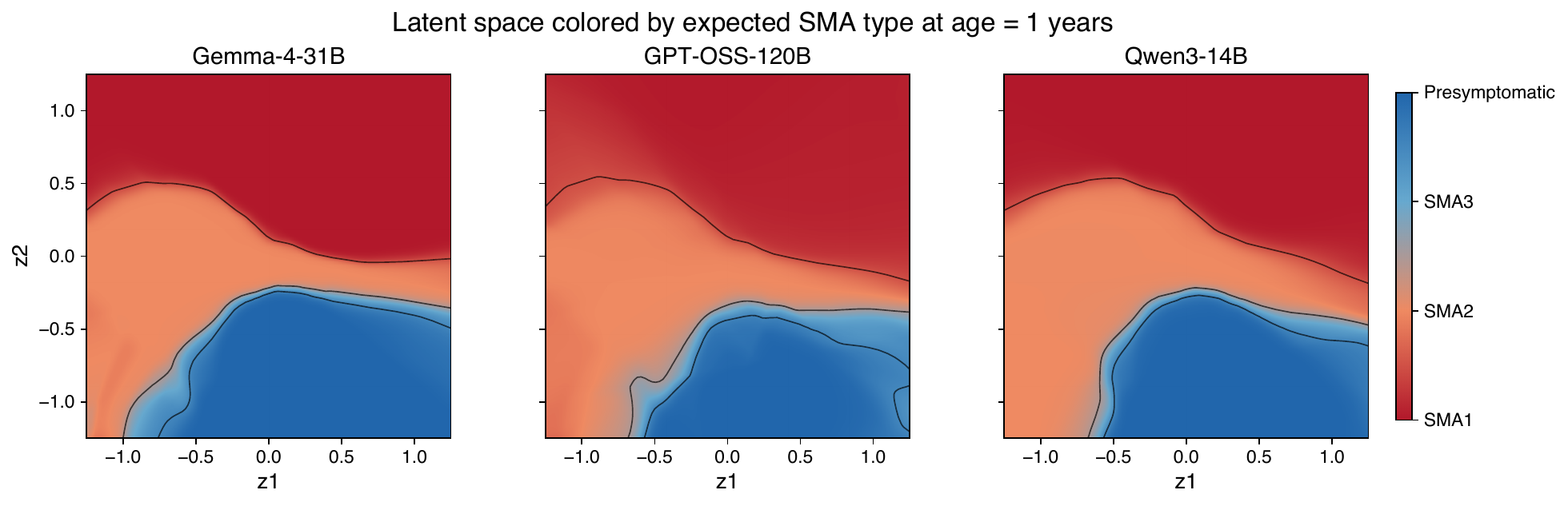}
        \caption{Latent-space regions colored by surrogate-predicted SMA type.}
    \end{subfigure}
    \caption{Interpretability of the synthetic-expert surrogate in profile space and latent space. 
    The upper panel shows age-dependent SMA type probabilities for three patient groups: sitting without standing (observations with HINE-2 sitting item score = 4, standing < 2), standing without walking (standing = 3, walking < 2), and walking (walking = 3, corresponds typically with a full HINE-2 sum score with no recorded deficiencies). 
    Colors indicate SMA type labels and marker shapes indicate the LLM model.\\
    The lower panel visualizes the latent space at age = 1 year colored by different synthetic experts. Each point in the latent grid is decoded into a motor-function profile and then classified by the selected LLM-specific surrogate network.}
    \label{fig:expert_interpretability}

    \par\smallskip
    \noindent\textbf{Alt text:}
    Composite figure with two rows of three panels. Panel (a) shows predicted
    probabilities for four SMA-related categories as functions of age for
    profiles characterized by sitting without standing, standing without
    walking, and independent walking. Curves compare the Gemma-4-31B,
    GPT-OSS-120B, and Qwen-3-14B surrogate models. Sitting-only profiles
    generally shift from presymptomatic at very young ages toward SMA type 2;
    standing profiles shift toward SMA types 2 or 3; and walking profiles
    remain predominantly presymptomatic. Panel (b) shows category regions
    across two latent dimensions at age one year for the same three models.
    All models produce broadly similar severity-ordered regions, although
    their category boundaries differ locally.
    
\end{figure}

The surrogate provides a fast, differentiable approximation to the input-output mapping induced by the selected LLM, prompt, decoding procedure, and aggregation rule. 
To visualize this, we evaluated the decision boundaries for the sit, stand and walk motor-function items. 
Here we grouped observations into motor function profiles that achieved independent sitting but not standing, independent standing but not walking, and walking. 
We varied the age values and averaged the predicted class probabilities over all samples of the dataset. 
We show these results in Figure~\ref{fig:expert_interpretability}.
We observed that the same motor-function profile receives different class probabilities depending on age, showing that the surrogate captures the developmental context used by the synthetic expert models.
Consistent with the clinical expectation, motor function profiles of patients that have sitting as highest motor function are eventually classified as SMA type 2 after being labeled presymptomatic at a young age.
Patients who can stand but not walk are classified as either SMA type 2 or SMA type 3, whereas patients who can walk are predominantly classified as presymptomatic. 
The latter group typically has no recorded motor deficits on the HINE-2 because these patients reach the upper limit of the scale.
We also use the synthetic expert to make latent variables interpretable in terms of SMA type categories, by decoding latent variables and classifying the generated samples by the surrogate network.
This allows us to assess latent decision boundaries. We show these results in Figure~\ref{fig:expert_interpretability}. 

\subsection{Consequences for consistency and reconstruction loss}

We next evaluated how synthetic-expert supervision affected the reconstruction behavior of the cVAE-based model.
For this analysis, we used surrogate networks derived from the most consistent LLMs within each evaluated model family.

Specifically, we tracked the Jensen-Shannon divergence between the categorical surrogate judgment 
distributions and the proportion of profiles for which the observed and 
reconstructed profiles received different expected SMA type labels from the surrogate model.
To assess whether improvements in SMA-type label consistency came at the cost of numerical reconstruction quality, 
we additionally recorded the average HINE-2 sum score reconstruction error.
Finally, we also evaluated the reconstruction error for profiles with initial label disagreement under the unsupervised baseline.
Analyses were repeated over 10 random seeds and across different choices of supervision weights $\lambda_{\mathrm{SE}}\in\{0,10,25,50,100\}$.
The resulting supervision sweep is summarized in Table~\ref{tab:llm_consistency_reconstruction}.

Across the LLM model families, increasing the supervision weight reduced the discrepancy between observed profiles and their reconstructions in the surrogate judgment space.
We observed this pattern both for the distributional comparison of class probabilities via JS divergence and for the total error rate in expected SMA type labels.
Therefore, the supervision signal showed a consistent shift toward reconstructions that retained the same clinical interpretation by the surrogate model as the corresponding input profiles.

This gain in consistency with the LLM-derived functional categorization was not necessarily accompanied by a substantial trade-off in reconstruction accuracy.
Across the tested supervision weights, the average HINE-2 reconstruction error remained close to the unsupervised baseline, with small deterioration in reconstruction quality for larger supervision weights.
However, profiles whose unsupervised reconstructions crossed an SMA type boundary under the unsupervised cVAE baseline saw a noticeable increase in reconstruction quality.  
The supervision term seemed to shift the model capacity toward samples whose reconstructions cross a clinical label boundary. 
As a consequence, samples without label disagreement received less emphasis in the fitted objective, leading to a small deterioration in their numerical reconstruction quality.

\subsection{Milestone prediction with Cox regression}

Finally, we used a Cox regression to assess whether the supervised latent representations carried prognostic information for future functional milestones.
The SMArtCARE dataset tracks sitting, standing, and walking milestones, which are related to the corresponding HINE-2 items but not identical to them.
For each milestone, visits were represented as time-varying Cox intervals on the age scale.
The fitted models predicted milestone attainment within one year after each held-out landmark visit.

Performance was compared with baselines based on the HINE-2 sum score, the HINE-2 sum score plus fixed clinical covariates, and the latent representation of the cVAE base model.
Supervised latent representations were evaluated across synthetic-expert model families and supervision weights (Table~\ref{tab:cox_llm_milestones}).
We evaluated model performance using three metrics. 
First, we used the IPCW Brier score to quantify the pointwise accuracy of predicted one-year milestone risk. 
Second, we used the time-dependent AUC to assess discrimination between patients who attained the milestone within one year and those who did not. 
Third, we evaluated calibration by computing the weighted mean absolute difference between predicted and observed one-year risk across calibration bins. 
All metrics were averaged over 10 random seeds and computed with patient-grouped five-fold cross-validation.

We observed that the Cox models fitted on the latent representation of the cVAE base model outperformed the data-level baselines on the reported metrics. 
The supervised variants generally preserved this prognostic information and further improved performance relative to the unsupervised latent baseline. 
This pattern is consistent with the role of the synthetic-expert signal: because SMA type categories are partly defined by expected milestone attainment, 
a supervision term that encourages clinically consistent SMA type interpretation may also encourage the latent representation to retain information relevant for milestone prediction.
However, the supervision weight should be chosen carefully; for larger values like $\lambda_{\mathrm{SE}}=100$ we observed a decrease in performance on some metrics, whereas moderate weighting like $\lambda_{\mathrm{SE}}=25$ appeared to provide a balance between supervision signal and the original cVAE loss terms.

\section{Discussion}
\label{discussion}

In this work, we proposed a framework for using LLM-derived clinical judgments as a supervision signal for longitudinal rare-disease modeling.
We were motivated by the difficulty of incorporating clinical knowledge into model fitting, particularly when relevant distinctions are well understood by experts and described in the medical literature, but are difficult to translate into explicit formulas, priors, or constraints.
To address this, we proposed an approach that uses an LLM as a synthetic clinical expert to assign clinical labels to textual descriptions of observations. 
These labels are then used to train a differentiable surrogate, which provides a supervision signal for steering the training of a generative representation model. 
Our approach targets reconstructions that remain close to their input in terms of numerical reconstruction error, but change the clinical interpretation of the original profile.

We evaluated the approach on spinal muscular atrophy motor function data, where we extracted clinical judgments with a pairwise comparison prompt design.
We found consistency of these judgments to be highly dependent on both model size and model family.
Since the computational cost of querying a more capable model comes only during offline label generation, 
the LLM quality should be prioritized before its outputs are used as part of a statistical loss.

When our approach was applied to steer a cVAE-based approach we observed an improvement in consistency under the LLM-derived surrogate via a reduced disagreement between its judgment distributions of observed and reconstructed profiles.
For moderate supervision weights HINE-2 reconstruction errors remained in a similar range, indicating that the additional loss component did not hinder the numerical reconstruction objective.
However, the relevance of synthetic-expert supervision depends on whether the surrogate has learned clinically meaningful distinctions, which in turn depends on the quality of the LLM-derived judgments.
We therefore used a Cox regression as a complementary analysis to investigate whether supervision also improves encoding of information relevant for future motor milestone attainment.
The milestone prediction results were supportive in this respect. Because SMA type categories are closely tied to expected attainment of sitting, standing, and walking milestones, 
improved consistency with respect to SMA type interpretation also encouraged the latent representation to more accurately encode milestone-relevant information.

However, several methodological limitations follow from our approach.
First, the synthetic expert is only as appropriate as the prompt, label definitions, textual rendering, and LLM used to generate the judgments.
If systematic errors in the synthetic expert are distilled into the surrogate, they become part of the training objective and shape the learned latent representation.
For this reason, synthetic-expert outputs should be carefully audited before deploying the approach in a clinical setting.
We used consistency checks to identify internal contradictions, sensitivity to label order, and unstable answer patterns.
However, these checks cannot prove that every retained label is clinically correct.
Moreover, uncertainty of the synthetic expert is currently used mainly as a diagnostic property. 
Future extensions could therefore propagate uncertainty into the loss, for example by down-weighting uncertain labels or using softer target distributions.

A second limitation is that the LLM-derived judgment is not used directly during cVAE training, but only through a learned differentiable surrogate.
This is necessary because it makes supervision computationally feasible and provides gradients, but it also introduces an approximation error.
Especially in sparse or ambiguous regions of the HINE-2 profile space, the surrogate may smooth over uncertainty and fail to reproduce exact decision boundaries of the original synthetic expert.
This could be mitigated by using uncertainty-aware surrogate models or re-querying the LLM for profiles near clinical decision boundaries.

A third limitation concerns the supervision weight.
Synthetic-expert supervision acts as a regularization term, and its effect depends on the balance between clinical consistency and numerical reconstruction.
Low weights may be too weak to influence the learned representation, whereas high weights may overemphasize the synthetic label space and reduce reconstruction quality and even downstream prediction performance.
The supervision weight should therefore be selected as an additional model hyperparameter, ideally using validation criteria that reflect both numerical reconstruction and the intended clinical use of the latent representation.

Taken together, our results support synthetic-expert supervision as a mechanism for incorporating clinical judgment into the disease progression modeling process.
Our broader vision is not to replace clinical expertise, but to create a more practical interface between clinical reasoning and statistical modeling.
Toward this goal, future work should strengthen clinical validation and propagate uncertainty from the synthetic expert into the loss.
If these steps are successful, language-based clinical judgment and probabilistic modeling could become connected parts of a single workflow in rare-disease research.

\section*{Funding}

This work was supported by the Deutsche Forschungsgemeinschaft (DFG, German Research Foundation) [Project-ID 499552394 -- SFB 1597 Small Data].

\section*{Ethics statement} 
The use of anonymized data from the SMArtCARE registry was reviewed by the Ethics Committee of the University of Freiburg (reference no. 22-1430-S1-retro). The SMArtCARE registry authorized provision of the anonymized data for this research project.
To protect participant confidentiality, all LLM inference was performed locally within secure computing environments.

\section*{Data and code availability}
The SMArtCARE data cannot be made publicly available because of participant privacy restrictions. The analysis code will be made publicly available upon publication and will be provided to reviewers upon request.

\section*{Conflicts of interest}

Harald Binder serves as a Guest Editor of the \textit{Biostatistics}
special collection \textit{Statistical Foundations of AI and Real-World
Evidence Generation}. The remaining authors declare no conflicts of interest.

\section*{Acknowledgments}

Generative AI tools (ChatGPT, using GPT-5.4 and GPT-5.5) were used for language editing and for reviewing and debugging code written by the authors. No patient-level data were transmitted to these services. All AI-assisted text and code changes were reviewed, verified, and revised by the authors, who take full responsibility for the final manuscript and code.

\printbibliography

\begin{table}[!p]
    \centering
    \scriptsize
    \setlength{\tabcolsep}{5pt}
    \caption{Diagnostics of LLM-derived synthetic-expert judgments and surrogate training. 
    Ordinal consistency is the percentage of profiles with internally consistent pairwise answers. 
    Pairwise and final-label agreement are measured under label-order reversal. Final-label agreement is defined as when the final label agrees under both prompt presentation orders, or the pairwise questions completely align for inconsistent answer patterns.
    Surrogate accuracy is the cross-validated accuracy for reproducing retained synthetic-expert labels. Values for the coin toss baseline are analytical values.
    The uncertainty ratio is the mean self-reported uncertainty for pairwise answers that changed under label-order reversal, divided by the mean for answers that remained stable. Here, low was mapped to an uncertainty score of 0, medium to 1 and high to 2.}
    \label{tab:llm_output_quality}
    \begin{tabular}{lccccc}
    \toprule
    Synthetic expert & \makecell{Ordinal\\consistency (\%)} & \makecell{Rev. pairwise\\agreement (\%)} & \makecell{Rev. final-label\\agreement (\%)} & \makecell{Surrogate\\accuracy (\%)} & \makecell{Uncertainty\\ratio} \\
    \midrule
    Coin Toss       & 9.4   & 50.0  & 1.7   & 33.3   & -- \\
    \midrule
    Gemma-4-E2B     & 74.1  & 85.3  & 52.8  & 89.6   & 1.5 \\
    Gemma-4-E4B     & 80.0  & 84.6  & 47.8  & 84.8   & 1.3 \\  
    Gemma-4-12B     & 87.1  & 95.7  & 85.1  & 94.9   & \textbf{11.4} \\    
    Gemma-4-31B     & \textbf{89.9}  & \textbf{97.8}  & \textbf{92.7}  & \textbf{97.2}   & 9.6  \\
    \midrule
    GPT-OSS-20B     & 74.1  & 85.8  & 48.7  & 87.5   & 1.6 \\
    GPT-OSS-120B    & 82.0  & 90.2  & 63.5  & 91.8   & 2.2 \\
    \midrule
    Qwen-3-0.6B     & 10.9   & 58.1  & 6.2   & 63.4   & 1.0  \\
    Qwen-3-1.7B     & 34.7  & 68.6  & 14.9  & 68.7   & 1.0  \\
    Qwen-3-4B       & 48.6  & 83.8  & 42.6  & 91.1   & 1.4  \\
    Qwen-3-8B       & 80.7  & 91.8  & 70.0  & 96.1   & 2.1  \\
    Qwen-3-14B      & 82.1  & 93.6  & 76.5  & 92.4   & 2.2  \\
    \bottomrule
    \end{tabular}
 \end{table}

\begin{table}[!p]
    \centering
    \scriptsize
    \setlength{\tabcolsep}{5pt}
    \caption{Effect of synthetic-expert supervision weight on reconstruction consistency and reconstruction error. Lower values indicate better consistency or lower reconstruction error. Label disagreement is averaged over the forward and reverse label presentation orders. Bold values indicate the best value within each synthetic-expert model class and metric.}
    \label{tab:llm_consistency_reconstruction}
    \begin{tabular}{lccccc}
    \toprule
    Synthetic expert 
    & $\lambda_{\mathrm{SE}}$
    & JS divergence
    & \makecell{Label\\disagreement (\%)}
    & \makecell{HINE-2 sum-score\\error}
    & \makecell{HINE-2 error\\among mismatches} \\
    \midrule
    Gemma-4-31B & 0   & 0.054 & 10.9 & 1.359 & 2.451 \\
                & 10  & 0.043 & 9.0  & \textbf{1.357} & 2.012 \\
                & 25  & 0.038 & 8.2  & 1.360 & 2.010 \\
                & 50  & 0.034 & 7.5  & 1.377 & 2.004 \\
                & 100 & \textbf{0.031} & \textbf{6.8}  & 1.382 & \textbf{2.001} \\
    \midrule
    GPT-OSS-120B & 0   & 0.034 & 10.1 & 1.359 & 2.354 \\
                 & 10  & 0.029 & 8.7  & 1.367 & 1.924 \\
                 & 25  & 0.026 & 8.1  & \textbf{1.358} & 1.910 \\
                 & 50  & 0.024 & 7.7  & 1.361 & 1.889 \\
                 & 100 & \textbf{0.022} & \textbf{7.2}  & 1.370 & \textbf{1.872} \\
    \midrule
    Qwen-3-14B & 0   & 0.032 & 8.9 & 1.359 & 2.351 \\
               & 10  & 0.027 & 7.9 & 1.356 & 2.005 \\
               & 25  & 0.025 & 7.3 & \textbf{1.345} & 1.898 \\
               & 50  & 0.022 & 6.7 & 1.354 & \textbf{1.827} \\
               & 100 & \textbf{0.021} & \textbf{6.4} & 1.425 & 1.853 \\
    \bottomrule
    \end{tabular}
\end{table}

\begin{table}[!p]
    \centering
    \scriptsize
    \setlength{\tabcolsep}{4pt}
    \caption{Milestone prediction on latent representations influenced by synthetic-expert supervision. 
    Rows report one-year landmark prediction performance for a linear time-varying Cox model. 
    Brier denotes the inverse-probability-of-censoring weighted Brier score, 
    AUC the corresponding time-dependent AUC, 
    and ECE the weighted expected calibration error across calibration bins. 
    Lower Brier and ECE values and higher AUC values indicate better performance. 
    Bold values indicate the best performance across all latent-representation feature sets for each milestone-metric combination.
    }

    \label{tab:cox_llm_milestones}
    \resizebox{\textwidth}{!}{%
    \begin{tabular}{llrrrrrrrrr}
    \toprule
    & & \multicolumn{3}{c}{Sitting} & \multicolumn{3}{c}{Standing} & \multicolumn{3}{c}{Walking} \\
    \cmidrule(lr){3-5}\cmidrule(lr){6-8}\cmidrule(lr){9-11}
    Feature set
    & $\lambda_{\mathrm{SE}}$
    & Brier & AUC & ECE
    & Brier & AUC & ECE
    & Brier & AUC & ECE \\
    \midrule
    HINE-2 sum score & -- & 0.271 & 0.817 & 0.314 & 0.180 & 0.834 & 0.175 & 0.142 & 0.845 & 0.134\\
    HINE-2 sum score + covariates & -- & 0.209 & 0.890 & 0.255 & 0.155 & 0.908 & 0.153 & 0.124 & 0.919 & 0.119\\
    \midrule
    $\mathbf{Z}$ & 0 & 0.097 & 0.944 & 0.078 & 0.098 & 0.951 & 0.102 & 0.087 & 0.954 & 0.088\\
    \midrule
    \multirow{4}{*}{\makecell[l]{$\mathbf{Z}$ informed by\\\hspace{1em}Gemma-4-31B}} & 10  & 0.085 & 0.954 & 0.067 & 0.084 & 0.962 & 0.076 & 0.067 & 0.964 & 0.065\\
                & 25  & 0.084 & 0.956 & 0.064 & 0.088 & 0.960 & 0.081 & 0.071 & 0.962 & 0.067\\
                & 50  & 0.085 & 0.952 & 0.058 & 0.089 & 0.956 & 0.079 & 0.070 & 0.962 & 0.066\\
                & 100 & 0.091 & 0.951 & 0.075 & 0.094 & 0.954 & 0.084 & 0.074 & 0.960 & 0.068\\
    \midrule
    \multirow{4}{*}{\makecell[l]{$\mathbf{Z}$ informed by\\\hspace{1em}GPT-OSS-120B}} & 10  & 0.086 & 0.952 & 0.061 & \textbf{0.079} & 0.962 & \textbf{0.063} & \textbf{0.064} & \textbf{0.967} & 0.065\\
                 & 25  & \textbf{0.083} & 0.954 & \textbf{0.057} & 0.082 & \textbf{0.964} & 0.073 & 0.066 & \textbf{0.967} & \textbf{0.063}\\
                 & 50  & 0.091 & \textbf{0.958} & 0.090 & 0.096 & 0.959 & 0.092 & 0.077 & 0.959 & 0.065\\
                 & 100 & 0.098 & 0.955 & 0.108 & 0.096 & 0.956 & 0.087 & 0.076 & 0.957 & 0.072\\
    \midrule
    \multirow{4}{*}{\makecell[l]{$\mathbf{Z}$ informed by\\\hspace{1em}Qwen-3-14B}} & 10  & 0.094 & 0.951 & 0.070 & 0.089 & 0.962 & 0.096 & 0.078 & \textbf{0.967} & 0.080\\
               & 25  & 0.091 & 0.952 & 0.072 & 0.089 & 0.958 & 0.082 & 0.075 & 0.958 & 0.071\\
               & 50  & 0.096 & 0.949 & 0.071 & 0.091 & 0.958 & 0.091 & 0.083 & 0.961 & 0.084\\
               & 100 & 0.102 & 0.939 & 0.077 & 0.097 & 0.955 & 0.096 & 0.081 & 0.959 & 0.087\\
    \bottomrule
    \end{tabular}
    }
    \end{table}

\end{document}


\maketitle

\setcounter{table}{0}
\renewcommand{\thetable}{S\arabic{table}}

\appendix

\section{Data preprocessing and hyperparameters}
\label{hyperparameters}

\subsection{Dataset}

The analysis used HINE-2 motor-function assessments of patients under the age of 12 years from the SMArtCARE registry. 
The SMArtCARE registry tracks the motor function development, motor milestone achievement and treatment history of children diagnosed with spinal muscular atrophy.
The HINE-2 profile contains eleven motor-function items regarding upper limb function, raising hands, reaching overhead, head control, sitting, voluntary grasp, kicking, rolling, crawling, standing, and walking. 
The measured motor abilities of these categorical item responses were mapped to ordered numerical levels, and visits with missing values were excluded.
For neural-network inputs, each ordinal HINE-2 item was subsequently thermometer-encoded.
To retain longitudinal information for latent mixed-effects modeling, patients were required to have more than four complete HINE-2 visits.

In total the dataset contained exactly 13,000 observations and 994 patients, with a median of 12 visits per patient. 
The median age at first observation was 1.2 years, and the median follow-up duration was 3.5 years.

\subsection{Synthetic-expert label generation and surrogate training}

To generate a training dataset for the surrogate network we supplemented the real HINE-2 data with synthetic samples. 
For this we fitted a two-dimensional age-conditional VAE model to the HINE-2 profiles. 
Subsequently, synthetic profiles were sampled from the decoder until a total of 20,000 unique generated profiles were retained. 
To allow for slight variations in ages, the age values were jittered by up to 0.25 years. 

Each profile was rendered as a textual motor-function description together with the age value. 
The item-level wording was based on the HINE-2 item descriptions \cite{HINE}. 
The exact score-to-text conversion used for this rendering is shown in Table~\ref{tab:hine_text_rendering}.

{\scriptsize
\setlength{\LTpre}{4pt}
\setlength{\LTpost}{4pt}
\setlength{\tabcolsep}{3pt}
\setlength{\parskip}{1pt}
\renewcommand{\arraystretch}{0.9}
\begin{longtable}{p{1.6cm}p{12.9cm}}
\caption{Text rendering of HINE-2 profile scores used for the synthetic-expert prompts. Each generated profile was converted by selecting the sentence corresponding to the observed item score and concatenating the resulting item-level descriptions with the patient's age.}
\label{tab:hine_text_rendering}\\
\toprule
HINE-2 item & Text fragment by item score \\
\midrule
\endfirsthead
\caption[]{Text rendering of HINE-2 profile scores used for the synthetic-expert prompts (continued).}\\
\toprule
HINE-2 item & Text fragment by item score \\
\midrule
\endhead
\midrule
\multicolumn{2}{r}{Continued on next page} \\
\endfoot
\bottomrule
\endlastfoot
Introduction & The child is \texttt{\{years\}} years \texttt{\{months\}} months old when its motor function is evaluated. The child has the following motor function skills: \\
\midrule
Upper limb function & \textbf{0}: Cannot achieve useful hand function to grab objects.\par \textbf{1}: Achieves useful hand function to grab objects. \\
\midrule
Raise hands & \textbf{0}: Cannot raise its hands to the mouth.\par \textbf{1}: Raises its hands to the mouth. \\
\midrule
Reach overhead & \textbf{0}: Cannot reach overhead.\par \textbf{1}: Reaches overhead. \\
\midrule
Head control & \textbf{0}: Has no stable head control and cannot maintain the head upright.\par \textbf{1}: Holds the head upright briefly, but the head wobbles and control is unsteady.\par \textbf{2}: Maintains the head upright consistently and does not show head wobbling. \\
\midrule
Sitting & \textbf{0}: Cannot maintain a sitting position in a meaningful way even with external support.\par \textbf{1}: Sits only if the examiner stabilizes the pelvis or hips and cannot sit independently.\par \textbf{2}: Sits with propping and uses the arms or hands for balance, but cannot sit freely without upper limb support.\par \textbf{3}: Sits independently without obvious propping, but cannot rotate or pivot while sitting.\par \textbf{4}: Sits independently and rotates and pivots while sitting. \\
\midrule
Voluntary grasp & \textbf{0}: Does not show a voluntary grasp and cannot intentionally take or hold an object.\par \textbf{1}: Grasps using the whole hand, but cannot perform a more refined finger-thumb grasp or pincer grasp.\par \textbf{2}: Grasps using index finger and thumb in an immature grasp, but cannot perform a pincer grasp.\par \textbf{3}: Performs a pincer grasp and picks up objects with a mature finger-thumb grip. \\
\midrule
Kicking & \textbf{0}: Cannot kick in the supine position.\par \textbf{1}: Kicks horizontally in the supine position, but cannot lift the legs upwards.\par \textbf{2}: Kicks upward vertically in the supine position, but cannot reach the legs with the hands.\par \textbf{3}: Lifts the legs and touches the leg in the supine position, but cannot reach the toes.\par \textbf{4}: Lifts the legs in the supine position high enough to touch the toes. \\
\midrule
Rolling & \textbf{0}: Cannot roll onto the side from supine.\par \textbf{1}: Rolls onto the side from supine, but cannot complete rolling between supine and prone.\par \textbf{2}: Rolls from prone to side and supine, but cannot roll from supine to prone.\par \textbf{3}: Rolls fully in both directions, including from supine to prone and vice versa. \\
\midrule
Crawling & \textbf{0}: Cannot lift the head in a way that supports crawling.\par \textbf{1}: Supports weight on the elbows, but cannot push up on outstretched hands or crawl.\par \textbf{2}: Supports weight on outstretched hands, but cannot crawl forward.\par \textbf{3}: Crawls flat on the abdomen, but cannot crawl on hands and knees.\par \textbf{4}: Crawls on hands and knees. \\
\midrule
Standing & \textbf{0}: Cannot bear weight through the legs when held upright.\par \textbf{1}: Bears weight through the legs when held upright, but cannot stand with external support.\par \textbf{2}: Stands with external support, but not unaided.\par \textbf{3}: Stands upright, stable and unaided. \\
\midrule
Walking & \textbf{0}: Cannot show walking-related motor behavior and cannot bounce, cruise, or walk independently.\par \textbf{1}: Shows early supported stepping behavior, but cannot cruise along furniture or walk independently.\par \textbf{2}: Cruises while holding on for support, but cannot walk independently.\par \textbf{3}: Walks independently for at least several seconds. \\
\end{longtable}
}

Each profile was then queried through pairwise comparisons between the candidate labels SMA type 1, SMA type 2, SMA type 3, and presymptomatic motor development. 
Additionally, the pairwise comparisons were repeated under label-order reversal, resulting in two comparison panels per profile. 
The prompt template used for each pairwise comparison is shown in Table~\ref{tab:pairwise_prompt_template}.

\begin{table}[H]
\centering
\scriptsize
\caption{Prompt template for one pairwise synthetic-expert comparison. The placeholders \texttt{\{label A\}}, \texttt{\{label B\}}, and \texttt{\{profile text\}} were filled for each candidate-label pair and generated HINE-2 profile.}
\label{tab:pairwise_prompt_template}
\begin{tabularx}{\textwidth}{lX}
\toprule
Prompt part & Text \\
\midrule
Comparison instruction & Decide between two candidate motor-function labels for a child being evaluated for spinal muscular atrophy (SMA): \texttt{\{label A\}} or \texttt{\{label B\}}. \\
Decision rule & Use the child's age to contextualize the observed motor abilities. If neither label is a reasonable match, choose the allowed label whose severity is closer to the observed motor impairment. **When \texttt{\{label A\}} or \texttt{\{label B\}} is presymptomatic the following additional instruction was included:** Choose presymptomatic when motor abilities are age-appropriate and no SMA-related motor impairment is observed. \\
Profile text & \texttt{\{profile text\}} \\
Output constraint & Think briefly about your answer. Report your confidence of your final decision and conclude by returning the following JSON object: \texttt{\{"label": "{label B}", "certainty": "high|medium|low"\}} or \texttt{\{"label":"{label A}", "certainty": "high|medium|low"\}}. \\
\bottomrule
\end{tabularx}
\end{table}

The pairwise answers were aggregated into final SMA type labels by using the ordinal structure of the four labels,
SMA type 1 $<$ SMA type 2 $<$ SMA type 3 $<$ presymptomatic motor development.
Response patterns containing cycles or non-monotone preferences were classified as internally contradictory.
Profiles for which the label order failed to produce a valid final label were excluded from surrogate training.
The differentiable surrogate was then trained as a classifier to reproduce the retained labels from patient age and the thermometer-encoded HINE-2 profile.
Training was performed with two heads to reproduce the answer for each label order. Results were averaged over both heads. 

\subsection{Hyperparameters}

All neural network architectures were chosen to contain a single hidden layer with 100 units and $\mathrm{ReLU}$ activation function. 
Thermometer encoded HINE-2 profiles together with the age value amounted to 41 input units to the encoder network and surrogate model.
The surrogate was trained with cross-entropy loss on class labels, using the Adam optimizer with default hyperparameters and 50 training epochs.

The conditional VAE used a two-dimensional latent space and received age as conditioning information. 
The decoder parameterized an ordinal categorical likelihood for the HINE-2 items. 
cVAE parameters were optimized with Adam a learning rate of 0.01 over 20 alternating training epochs. 
The KL term was weighted by $\beta=0.5$, the alignment term by $\eta=5$. 
The synthetic-expert supervision weight was varied over $\lambda_{\mathrm{SE}}\in\{0,10,25,50,100\}$.

The latent mixed-effects model was fitted in the same two-dimensional latent space. For determining fixed-effect parameters, we followed our previous work \cite{Schaechter2025latentRepresentations} and included fixed effects for
age at symptom onset, \textit{SMN2} copy-number, time since treatment for the three recorded SMA medications, time since switching to a different SMA medication, BMI, height, and interaction terms with age. 
Treatment-time variables were set to zero before the respective treatment or switch. 
The random-effects structure contained a patient-specific random intercept and random slope. 
The mixed-effects parameters were re-estimated after each cVAE epoch with L-BFGS, using a learning rate of 0.1.

\subsection{Milestones and Cox regression}

The SMArtCARE registry records patient-level attainment of independent sitting, standing, and walking as motor milestones.
These endpoints correspond to the same broad motor domains as the HINE-2 items, but use separate milestone definitions.
For the Cox analysis, exact milestone-attainment ages were treated as uncensored events.
If follow-up information was available but the exact attainment time was not observed, patients were right-censored at the recorded age.
If future milestone attainment was unknown, they were right-censored at the last motor-function visit without a recorded milestone.
Entries indicating that a milestone had already been achieved before observation, but without an exact age, were left-censored and excluded.
The milestone definitions and number of observed events are shown in Table~\ref{tab:milestone_definitions}. 

\begin{table}[H]
\centering
\scriptsize
\setlength{\tabcolsep}{4pt}
\renewcommand{\arraystretch}{0.95}
\caption{Definitions of the motor milestones used for Cox regression. Events denote exact milestone attainments contributing an event interval after excluding left-censored patient-milestone pairs and pairs without a motor-function visit before the event time.}
\label{tab:milestone_definitions}
\begin{tabularx}{\textwidth}{lrX}
\toprule
Milestone & Events & Definition \\
\midrule
Sitting & 432 & Patient sits up straight with the head erect for at least 10 seconds. Patient does not use arms or hands to balance the body or to support the position. \\
Standing & 304 & Patient stands in an upright position on both feet, not on the toes, with the back straight. The legs support 100\% of the patient's weight. There is no contact with an object or a person. Patient stands alone for at least 10 seconds. \\
Walking & 264 & Patient takes at least 5 steps independently in an upright position with the back straight. One leg moves forward while the other supports most of the body weight. There is no contact with a person or object. \\
\bottomrule
\end{tabularx}
\end{table}

For each patient and milestone, the longitudinal visits before the event or censoring time were converted into time-varying Cox intervals on the age scale.
Each interval started at an observed visit age and ended at the next visit age or at the milestone or censoring time, whichever occurred first.
The evaluated feature sets comprised the HINE-2 total score, the HINE-2 total score plus fixed clinical covariates and latent encoder coordinates.
For the synthetic-expert supervision sweep, linear Cox models were fitted to the latent representations across synthetic experts with different choices of supervision weights.
Models were fitted with patient-grouped five-fold cross-validation.

Predicted risks were evaluated at a fixed one-year landmark horizon from each held-out visit.
For this horizon, a visit was treated as a case if the milestone occurred within one year and as a control if the patient remained event-free beyond the horizon.
Visits censored before the one-year horizon were handled through inverse-probability-of-censoring weighting.
Evaluation rows were additionally weighted inversely by the number of landmark visits per patient and milestone, so that patients with more visits did not dominate the metric calculation.
We reported three one-year prediction metrics.
Discrimination was measured by the inverse-probability-of-censoring weighted time-dependent AUC. Prediction error was measured by the corresponding weighted Brier score.
Calibration was summarized by the expected calibration error, defined as the weighted mean absolute difference between predicted and observed risk across 10 quantile-based calibration bins.